\documentclass[letterpaper]{article} 
\usepackage{aaai2027}  
\usepackage[hyphens]{url}  
\usepackage{graphicx} 
\usepackage{natbib}  
\usepackage{caption} 
\usepackage{algorithm}
\usepackage{algorithmic}

\usepackage{newfloat}
\usepackage{listings}
\DeclareCaptionStyle{ruled}{labelfont=normalfont,labelsep=colon,strut=off} 
\floatstyle{ruled}
\newfloat{listing}{tb}{lst}{}
\floatname{listing}{Listing}

\usepackage{booktabs}

\usepackage{amsmath}
\usepackage{tabularx}
\usepackage{multirow}
\usepackage{subcaption}

\nocopyright
\title{Affect-Prototype Guided Fusion for Open-Vocabulary Incomplete Multi-modal Emotion Recognition}
\author{
    Yichi Zhang\equalcontrib\textsuperscript{\rm 1},
    Shenyue Wang\equalcontrib\textsuperscript{\rm 2},
    Jing Luo\textsuperscript{\rm 1},
    Chunyang Yu\corresponding\textsuperscript{\rm 2},
    Xinyu Yang\corresponding\textsuperscript{\rm 1}
}
\affiliations{
\textsuperscript{\rm 1}School of Computer Science and Technology, Xi'an Jiaotong University, Xi'an, China\\
\textsuperscript{\rm 2}OPPO Research Institute, Shanghai, China\\
datasonezyc@stu.xjtu.edu.cn,wangshenyue@whu.edu.cn,chinglo@xjtu.edu.cn,chunyang.yu@ucalgary.ca,yxyphd@mail.xjtu.edu.cn
}

\begin{document}

\maketitle

\begin{abstract}
Open-vocabulary multimodal emotion recognition (OV-MER) aims to generate open natural-language emotion labels from multimodal affective cues. In real-world scenarios, however, complete and synchronized modal data are difficult to obtain due to limitations of acquisition devices and user privacy constraints. Existing OV-MER methods are largely designed for full-modal inputs, and fail to perform effective feature fusion under modal missing conditions. Meanwhile, current fusion approaches designed for incomplete modalities mainly focus on fixed-label recognition context, and cannot satisfy the demand for fuse emotional cues guided with arbitrary emotion semantics in OV-MER context. To tackle these challenges, this paper proposes an Affect-Prototype-Conditioned Fusion (APCF) framework for incomplete open-vocabulary emotion recognition. As a candidate-free generative framework, APCF extends modal contribution learning to scenarios guided by arbitrary emotional semantics. Specifically, we construct an affect-prototype library to explicitly model multimodal contribution characteristics corresponding to diverse emotions, which provides dynamic constraints for modal fusion under different emotional semantic perspectives. Conditional retrieval and feature aggregation are conducted based on available modal features. The refined fused affective representations are then fed into an LLM decoder to produce open-vocabulary emotion labels. Experiments on the OV-MERD+ and MER-FG datasets demonstrate that APCF substantially outperforms state-of-the-art baselines.
\end{abstract}


\begin{figure*}[ht]
    \centering
    \includegraphics[width=.9\linewidth]{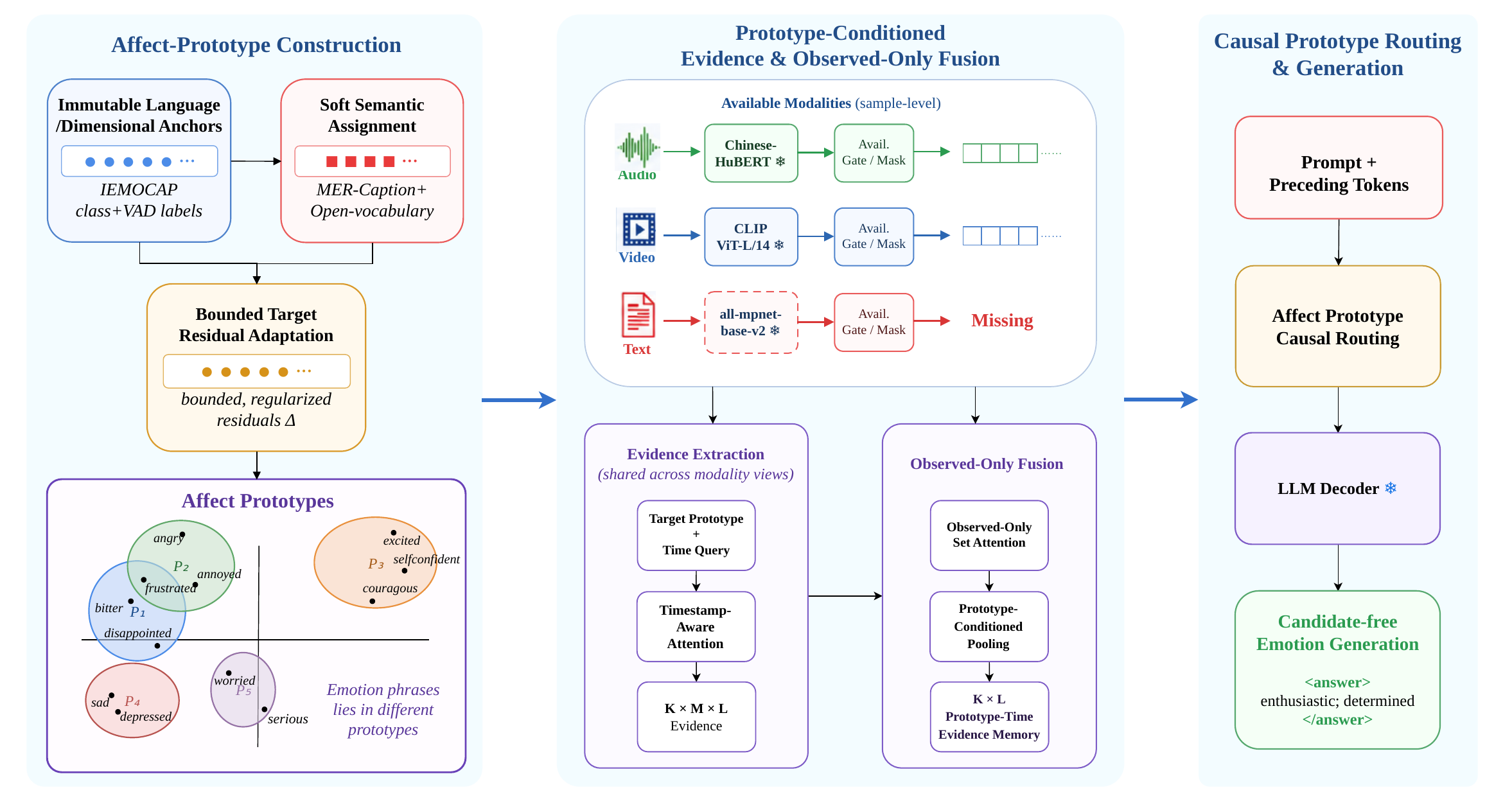}
    \caption{Overview of the proposed APCF framework.}
    \label{fig:main}
\end{figure*}

\section{Introduction}

Multimodal emotion recognition (MER) combines complementary audio, visual, and textual cues to infer a user's affective state. Most traditional approaches formulate MER as classification over a fixed inventory of emotion words as labels \cite{zhangTAILORVersatileMultimodal2022}. Although this formulation supports straightforward supervision and evaluation, a small label inventory compresses nuanced states and cannot express concepts outside the predefined classes. Continuous emotion recognition instead predicts values in dimensional emotion labels to express arbitrary affect states \cite{praveenJointCrossAttention2022}, however the dimension-based labels naturally present huge semantic gap between the human-understandable descriptions and dimensional descriptions, limiting dataset annotation and further usage on end-user facing systems. OV-MER proposed predicting emotion phrases in natural language for balancing semantic coverage with human readability \cite{lianOVMEROpenVocabularyMultimodal2025}. Such phrases can describe intensity, mixtures, and fine distinctions, but they also enlarge the semantic space that must be connected to sensory evidence. AffectGPT proved multi-modal large language models (MLLMs) is capable for generating fine-grained affect description based on multi-modal emotion evidence \cite{lianAffectGPTNewDataset2025}. Recent work further advances this interface through perception-oriented policy optimization \cite{hanOmniPerceptionPolicyOptimization2026} and reinforcement learning \cite{lianAffectGPTRLRevealingRoles2026}. Challenge-oriented generative evaluation \cite{ref05_mer2026} and hybrid-evidential deduction \cite{ref06_hydra} likewise support open-vocabulary emotion understanding.

Most current open-vocabulary MER pipelines nevertheless assume that all modalities are available. That assumption is fragile in practice: obtaining complete and synchronous modality data would be hard due to issues like sensor malfunction, data collection limitation and user privacy concerns. User may not look into the camera, or even reject giving permissions on capturing visual data. Transcriptions may not collect enough meaningful semantic information with heavily-accented or mainly-interjection speech. Incomplete modality data will hamper learning and reasoning abilities of the language modal on the relationship between affect phrase semantic and multi-modal emotional evidence. Lack of crucial trained emotion features, such as micro-expression, or the semantic information inside transcription, would impact emotion reasoning when fusing multi-modal emotional cues, and reduce the emotion recognition performance.

Incomplete-MER research has addressed missing inputs through reconstruction \cite{zhaoMissingModalityImagination2021}, learned prompts \cite{guoMultimodalPromptLearning2024}, and flexible expert routing \cite{hanFuseMoEMixtureofExpertsTransformers2024}. Other approaches use cross-modal queries \cite{miyoshiRobustMultimodalEmotion2026}, balanced prompting \cite{heCrossmodalPromptingBalanced2026}, or uncertainty-aware diffusion \cite{qiuMissingModalitiesHypergraph2026}. These methods improve the use of partial data for emotion understanding and reasoning. However, current algorithm design focused on semantic understanding under fixed-label context, not able to adapt its ability on learning emotional modality cues and fusion matrices to the arbitrary emotion space in open-vocabulary context. Suppose that a system is trained mainly with \textit{fearful} but is evaluated on \textit{apprehensive}. A language model may know that the phrases are related, yet the sensory pathway may not know that they should draw on a similar mixture of hesitant speech, tense facial behavior, and uncertain wording. As well as how to fully utilize the mixture of emotion cues to better instructing language models under incomplete multi-modal circumstances.

We propose \textbf{Affect-Prototype–Conditioned Fusion (APCF)} to address this problem. APCF maintains a set of overlapping landmarks in emotion-language space called affect-prototypes, which are aligned between categorical emotion word semantic features, and label coordinates in the dimensional emotion space. A prototype is a reusable query that conditions shared computation, it learns a range of similar emotion evidence-wise, and a evidence fusion algorithm for emotions inside the prototype. For each observed audio, video, and text stream, APCF extracts prototype-specific evidence and combines the available modality streams with an observed-set fusion module. The resulting prototype memory is exposed to a pretrained language decoder. Its causal prefix determines which semantic regions and which sample-specific evidence slots are relevant, instructing the decoder to fully reason and generate open-vocabulary emotion descriptions with incomplete modality cues.

Our main contributions are listed as follows:

\begin{enumerate}
    \item \textbf{We extend open-vocabulary MER to incomplete multi-modal scenario.} This paper formulates incomplete open-vocabulary MER as candidate-free generation from any nonempty subset of modalities, bridging open-vocabulary recognition and incomplete-modality fusion.
    \item \textbf{We construct semantic and modality-aware affect-prototypes.} Transferable prototypes are learned from paired categorical and dimensional source supervision, then adapted to target emotion phrases through bounded, regularized residuals. They organize both affect semantics and modality-specific evidence-use patterns.
    \item \textbf{We designed a novel modality fusion algorithm.} We use the prototypes to condition timestamped evidence retrieval, observed-only set fusion, and causal routing into a language encoder for reasoning and generating open-vocabulary emotional phrases. The algorithm achieves state-of-the-art (SOTA) on incomplete open-vocabulary MER task.
\end{enumerate}

\section{Related Work}

\subsection{Open-vocabulary Emotion Recognition}
OV-MER replaces closed emotion classes with natural-language terms and evaluates predictions through semantic normalization and an emotion-wheel representation \cite{lianOVMEROpenVocabularyMultimodal2025}. AffectGPT extends this formulation with emotion-oriented multi-modal instruction data and an audiovisual language-model scaffold \cite{lianAffectGPTNewDataset2025}. Emotion-LLaMA uses multi-modal instruction tuning to combine emotion recognition with explanation and reasoning \cite{chengEmotionLLaMAMultimodalEmotion2024}. Agent-MER applies hierarchical agent deliberation \cite{laiAgentMERCognitiveAgent2025}, Clue2Emo organizes multimodal clues before prediction \cite{zhangClue2EmoBrainInspiredFramework2026}, Nano-EmoX develops a compact multitask affect model \cite{ref19_nano_emox}, and OPPO optimizes multimodal emotion reasoning with a perception-oriented policy objective \cite{hanOmniPerceptionPolicyOptimization2026}. AffectGPT-RL studies reinforcement learning for open-vocabulary recognition \cite{lianAffectGPTRLRevealingRoles2026}, HyDRA performs structured clue-based deduction \cite{ref06_hydra}, AffectAgent coordinates retrieval-augmented agents \cite{ref20_affectagent}, and AffectVerse predicts latent audiovisual futures for affective reasoning \cite{ref21_affectverse}. Together, these methods strengthen open-ended perception and reasoning. Some include missing-input mechanisms, but missingness is not uniformly studied across controlled modality subsets.

\subsection{Emotion Recognition with Incomplete Modalities}
The Missing Modality Imagination Network (MMIN) reconstructs unavailable representations through cross-modal imagination and cycle consistency \cite{zhaoMissingModalityImagination2021}. The multi-modal prompt-learning method of Guo et al.\ associates learned prompts with missing-input patterns \cite{guoMultimodalPromptLearning2024}, while the query-based method of Miyoshi et al.\ combines unimodal and cross-modal evidence \cite{miyoshiRobustMultimodalEmotion2026}. BALM balances training under unequal missing rates \cite{nguyenBALMModelAgnosticFramework2026}, and SimMLM trains over nested modality subsets so that additional observations do not degrade predictions \cite{liSimMLMSimpleFramework2025}. FuseMoE and Flex-MoE support arbitrary modality combinations through mixture-of-experts routing \cite{hanFuseMoEMixtureofExpertsTransformers2024,yunFlexMoEModelingArbitrary2024}. ComP propagates cross-modal prompts and consensus information between modality branches to balance fixed-label predictions \cite{heCrossmodalPromptingBalanced2026}. HyperEF uses hypergraph-conditioned diffusion to recover latent missing features and models source- and decision-level uncertainty in fixed-label conversational emotion recognition \cite{qiuMissingModalitiesHypergraph2026}. These approaches establish strong mechanisms for partial inputs, but their fusion modules are optimized primarily through fixed-class decisions. Applying them to a generator does not by itself provide an emotion-semantic query that transfers evidence-use patterns to previously unseen phrases.

\subsection{Semantic Prototypes and Affect Representation}
Semantic information has long been used to organize representations across labels and modalities. UniBind uses language-enhanced prototypes to construct a shared multi-modal space \cite{lyuUniBindLLMAugmentedUnified2024}. The semantic prompting method of Pipoli et al.\ conditions visual recognition on available-modality scenarios \cite{ref27_scp}, and the label-semantic model of Gaonkar et al.\ uses label descriptions to guide attention and relations between emotional reactions \cite{ref28_label_semantics}. The label-agnostic embeddings of Buechel et al.\ bridge heterogeneous categorical inventories \cite{ref13_label_agnostic}. Park et al.\ map categorical emotions to dimensional coordinates \cite{ref14_dimensional}, while the affective-manifold method of Li et al.\ anchors discrete terms in continuous affective structure \cite{ref15_anchoring}.

\section{Methodology}
\subsection{Problem Formulation}
Let \(\mathcal M=\{A,V,T\}\) denote audio, video, and text. Sample \(i\) contains a nonempty observed subset \(S_i\subseteq\mathcal M\) and an availability vector \(a_i\) that records which streams are present. A frozen modality encoder \(E_m\) maps each observed stream to
\begin{equation}
H_i^m=\{(h_{i,m,t},\tau_{i,m,t})\}_{t=1}^{T_i^m},
\qquad m\in S_i,
\end{equation}
where \(h_{i,m,t}\) is an encoded token and \(\tau_{i,m,t}\) is its normalized position in the clip. Every valid token follows this timestamp-aware representation. A naturally pooled stream is represented by a single valid token rather than expanded into an artificial temporal sequence. If a modality is unavailable, its encoder output is not created and no zero or learned missing-modality token is inserted into the evidence set.

The target \(Y_i=(y_{i,1},\ldots,y_{i,N_i})\) is an ordinary tokenizer sequence containing one or more free-form emotion phrases. APCF is therefore not given an emotion-specific output vocabulary or a test-time candidate list. Given a task prompt \(Q_i\) and an evidence memory \(Z_i^{S_i}\) built only from the observed streams, prediction is factorized causally as
\begin{equation}
p(Y_i\mid X_i^{S_i})
=\prod_n p(y_{i,n}\mid y_{i,<n},Q_i,Z_i^{S_i}).
\end{equation}
This formulation separates two requirements. The sensory pathway must reorganize whatever evidence remains under \(S_i\), while the language pathway must express that evidence using unrestricted emotion language. The structure of APCF is shown in Figure~\ref{fig:main}. APCF connects the two through affect prototypes: overlapping semantic queries that organize evidence without becoming output classes. The model contains \(K\) prototypes and \(L\) reference-time slots. Throughout this section, \(W_*\) denotes learned projections, and all unavailable streams are excluded from the corresponding operations.

\subsection{Affect-Prototype Bank}
APCF represents affect using overlapping landmarks rather than mutually exclusive output classes. The source bank is constructed only from samples for which categorical and dimensional annotations are paired. Prototype slots are initialized from source emotion categories, but a sample may assign mass to several slots. Prototype \(P_k^S\) contains a language anchor \(c_{k,S}^L\), a dimensional center \(c_{k,S}^D\), and a robust bandwidth \(\sigma_{k,S}\). A frozen semantic encoder embeds the category description. The dimensional center is estimated from the paired annotations, and the bandwidth summarizes within-category dispersion.

For source category \(y\), let \(g_y\) be its normalized semantic embedding and \(\bar v_y\) its robust dimensional center. Language and dimensional similarity induce two distributions:
\begin{align}
\alpha_{y,k}^{L}
&=\operatorname{softmax}_{k}
\left(\frac{\cos(g_y,c_{k,S}^{L})}{\tau_L}\right),\\
\alpha_{y,k}^{D}
&=\operatorname{softmax}_{k}
\left(-\frac{\lVert \bar v_y-c_{k,S}^{D}\rVert^2}
{\tau_D\sigma_{k,S}^{2}}\right).
\end{align}
The temperatures \(\tau_L\) and \(\tau_D\) control how broadly a category overlaps neighboring prototypes. Their bridge target is
\begin{equation}
\alpha_y^S=\lambda_B\alpha_y^L+(1-\lambda_B)\alpha_y^D,
\end{equation}
where \(\lambda_B\) controls the relative contribution of language and dimensional geometry. Because both annotations refer to the same source examples and share the same prototype slots, the bridge aligns categorical meaning with continuous affect structure instead of treating them as unrelated auxiliary tasks.

The source bank remains immutable during target training. This preserves a stable semantic coordinate system and prevents a small number of unusual target phrases from rewriting the source geometry. Directly freezing that geometry, however, would assume that source categories and target open-vocabulary descriptions organize affect identically. APCF therefore derives a target bank through bounded residual adaptation:
\begin{align}
c_{k,T}^L
&=\operatorname{norm}\!\left(c_{k,S}^L+\rho_L\tanh\Delta_k^L\right),\\
c_{k,T}^D
&=\operatorname{clip}\!\left(c_{k,S}^D+\rho_D\tanh\Delta_k^D\right).
\end{align}
The radii \(\rho_L\) and \(\rho_D\) bound how far a target prototype can move, while normalization and clipping keep the two components in their valid spaces. Source bandwidths remain fixed so that adaptation changes prototype location rather than silently changing what counts as a broad or narrow affect region.

Target emotion phrases provide soft supervision for the language residual through their similarity to the adapted anchors. No dimensional target is fabricated when a target corpus does not supply one. Instead, the dimensional residual receives gradients from downstream generation and is constrained by prototype-preservation terms. These terms penalize excessive residual norms, distortion of pairwise source relations, reversal of the source dimensional ordering, and collapse between target prototypes. The result is a target-aware bank that may shift toward domain-specific emotion language while remaining tethered to the source affect structure.

\subsection{Source-Domain Sensory Initialization}
The prototype bank is also used to initialize the sensory pathway. For each paired source example, the bridge distribution \(\alpha_y^S\) supervises temporary prototype-assignment heads, while a temporary regression head predicts the dimensional annotation. These heads are attached to unimodal evidence and to the fused source representation. Their joint objective can be summarized as
\begin{equation}
\mathcal L_{\mathrm{src}}
=\mathcal L_{\mathrm{assign}}
+\lambda_{\mathrm{dim}}\mathcal L_{\mathrm{dim}}
+\lambda_{\mathrm{cross}}\mathcal L_{\mathrm{cross}},
\end{equation}
where \(\mathcal L_{\mathrm{cross}}\) encourages compatible evidence organization across observed modalities. This stage teaches the shared evidence encoder that different affect regions may rely on different modality relations. It is therefore more than prototype construction: source supervision initializes the perceptual extraction and fusion parameters as well.

The temporary prediction heads are discarded after source training. The learned evidence parameters are transferred to the target stage, whereas the source anchors remain fixed and the target residuals begin from the source geometry. This separation prevents the source task from imposing a fixed output inventory on open-vocabulary generation.

\subsection{Prototype-Conditioned Evidence Extraction}
Each adapted prototype is transformed into a route code \(c_k\) and a sensory-conditioning code \(d_k\):

\begin{equation}
c_k=W_r[c_{k,T}^L;c_{k,T}^D],
\qquad
d_k=W_s[s_k;c_k],
\end{equation}

where \(s_k\) is a learned slot identifier without explicit semantic initialization. The route code carries affect meaning to the decoder, while the sensory code conditions shared perceptual computation. Keeping the slot identifier separate from the semantic code allows the architecture to preserve stable memory positions without equating a prototype with an output class.

For reference position \(\bar\tau_\ell\), the pair \((k,\ell)\) forms a query that retrieves evidence from each observed modality:
\begin{align}
q_{k,\ell}&=p_\ell+W_qd_k,\\
e_{i,k,m,\ell}
&=\operatorname{Attn}\!\left(
q_{k,\ell},H_i^m;
b_m(\tau_{i,m}-\bar\tau_\ell)\right).
\end{align}
Here \(p_\ell\) identifies a learned reference-time slot and \(b_m\) is a modality-specific relative-time bias. Timestamp embeddings are also added to the attention values. Consequently, the same sensory tokens can yield different evidence when queried from different affect regions or temporal positions. For example, a short vocal hesitation may be highly relevant to one prototype but weak evidence for another.

All prototypes and modalities share the retrieval parameters; only their conditioning codes and modality embeddings differ. This parameter sharing allows a target phrase near a source prototype to reuse computation rather than requiring a newly trained detector. Since retrieval is evaluated only for \(m\in S_i\), the evidence tensor contains no representation derived from an unavailable channel.

\subsection{Observed-Only Set Fusion}
Retrieved evidence must be combined without assuming a fixed number or order of observed streams. Let \(\eta_m\) be a modality identity embedding. APCF first applies prototype-conditioned FiLM \cite{ref31_film}, then uses a shared Set Transformer \cite{ref32_set_transformer}:
\begin{align}
x_{i,k,m,\ell}
&=\operatorname{FiLM}\!\left(
\operatorname{LN}(W_e e+\eta_m+W_a a_i);d_k\right),\\
r_{i,k,\ell}
&=\operatorname{Pool}_{d_k,a_i}
\left(\operatorname{SetAttn}
\{x_{i,k,m,\ell}:m\in S_i\}\right).
\end{align}
The availability encoding \(a_i\) tells the shared module which observation state produced the set, while the modality identities distinguish evidence sources. FiLM lets the same sensory content be emphasized differently under different affect prototypes. Set attention then models relations among only the available streams, and prototype-conditioned pooling aggregates their contributions into \(r_{i,k,\ell}\).

This operation is permutation invariant with respect to the presentation order of modalities and accepts every nonempty observed subset. Missingness is not represented by a synthetic feature, and the model is not required to reconstruct a plausible but unverifiable hidden stream. Instead, it learns how the contribution of an observed stream changes with both the affect query and the current observation state.

The fused slots form a prototype-by-time memory
\begin{equation}
z_{i,k,\ell}=W_g r_{i,k,\ell}+\eta_k^P+\eta_\ell^R,
\end{equation}
where \(\eta_k^P\) and \(\eta_\ell^R\) are semantic-free prototype and reference-slot identifiers. Semantic-free identifiers preserve memory structure; affect meaning enters through the prototype codes that conditioned retrieval and fusion.

\begin{figure}
    \centering
    \includegraphics[width=0.75\linewidth]{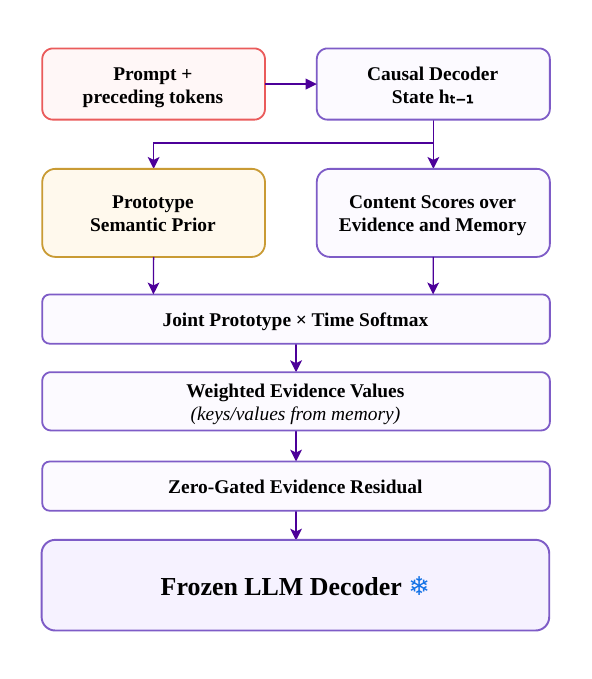}
    \caption{Structure of affect-prototype causal routing.}
    \label{fig:causal-routing}
\end{figure}

\subsection{Affect Prototype Causal Routing}
The structure of this module is detailed in Figure \ref{fig:causal-routing}. At a routed decoder layer, the causal state \(h_{i,n-1}\) contains only the prompt and previously generated tokens. Its projection \(q_{i,n}\) produces a semantic prior \(\beta\) over prototypes and content scores \(s\) over the sample-specific memory:

\begin{align}
\beta_{i,n,k}
&=\operatorname{softmax}_k
\left(\frac{\cos(q_{i,n},W_Pc_k)}{\tau_G}\right),\\
s_{i,n,k,\ell}
&=\frac{q_{i,n}^{\top}W_Kz_{i,k,\ell}}{\sqrt{d_h}},\\
a_{i,n,k,\ell}
&=\operatorname{softmax}_{k,\ell}
\left(s_{i,n,k,\ell}+\log(\beta_{i,n,k}+\epsilon)\right).
\end{align}

Here \(\tau_G\) is the routing temperature, \(d_h\) is the query dimension, and \(\epsilon\) is a numerical stabilizer. The semantic prior asks which affect regions are compatible with the current generation prefix; the content score asks which prototype-time slots contain relevant evidence for this particular sample. Their sum yields one joint distribution \(a_{i,n,k,\ell}\), so prototype choice and evidence choice cannot drift into independent decisions.

The routed evidence is

\begin{align}
v_{i,n}&=\sum_{k,\ell}a_{i,n,k,\ell}W_Vz_{i,k,\ell},\\
\widetilde h_{i,n}&=h_{i,n}+g\,W_Ov_{i,n},
\end{align}

where \(g\) is a zero-initialized residual gate. The value and output paths follow an identity-preserving factorization: learned transformations reconcile dimensions, while the prototype router controls selection rather than freely rewriting memory content. At initialization, the gate makes the augmented decoder equivalent to the frozen decoder. Training can then introduce evidence gradually without destabilizing its language ability.

Routing is strictly causal. A completed target phrase is never encoded and returned to the router during its own generation. At step \(n\), both the semantic prior and the content route depend only on \(Q_i\), \(y_{i,<n}\), and observed evidence. This prevents label leakage and permits ordinary autoregressive inference.

\subsection{Training Process}
Training proceeds from source prototype construction and sensory initialization to target-domain adaptation and candidate-free generation. The modality encoders and pretrained language decoder remain frozen; optimization updates the evidence modules, observed-set fusion, target prototype residuals, and causal router. At inference, target-phrase assignments, temporary source heads, and complete-view references are absent. APCF encodes only the observed streams, builds their prototype-conditioned memory, and generates emotion terms autoregressively.

\begin{table*}[t]
\centering
\small
\setlength{\tabcolsep}{3.5pt}
\caption{Open-vocabulary comparison on OV-MERD+ and MER-FG. Cells report (\%): level-1 F1 on OV-MERD+ and Avg on MER-FG.}
\label{tab:released-systems}
\begin{tabular}{lp{6cm}ccccccc}
\toprule
\multicolumn{2}{c}{} & \multicolumn{7}{c}{Modality view} \\
\cmidrule(lr){3-9}
Dataset & Method & A & V & T & AV & AT & VT & AVT\\
\midrule
\multirow{5}{*}{OV-MERD+}
& OV-MER \cite{lianOVMEROpenVocabularyMultimodal2025} & 33.914 & 45.211 & 47.085 & 44.969 & 48.304 & 50.163 & 46.952 \\
& Emotion-LLaMA \cite{chengEmotionLLaMAMultimodalEmotion2024} & 29.104 & 29.464 & 38.215 & 48.257 & 42.108 & 44.961 & 54.439 \\
& AffectGPT \cite{lianAffectGPTNewDataset2025} & 46.655 & 49.138 & 53.775 & 54.160 & 55.650 & 60.518 & 62.229 \\
& AffectAgent-R \cite{ref20_affectagent} & 48.226 & 49.941 & 53.594 & 55.502 & 58.173 & 61.004 & 62.832 \\
& \textbf{APCF (ours)} & \textbf{56.490} & \textbf{50.587} & \textbf{54.116} & \textbf{58.818} & \textbf{61.086} & \textbf{61.193} & \textbf{63.015} \\
\midrule
\multirow{5}{*}{MER-FG}
& OV-MER \cite{lianOVMEROpenVocabularyMultimodal2025} & 24.999 & 32.040 & 37.851 & 32.353 & 35.361 & 37.820 & 36.041 \\
& Emotion-LLaMA \cite{chengEmotionLLaMAMultimodalEmotion2024} & 17.533 & 20.976 & 26.024 & 31.410 & 28.699 & 29.471 & 36.889 \\
& AffectGPT \cite{lianAffectGPTNewDataset2025} & 33.925 & 31.132 & 34.904 & 35.598 & 36.379 & 39.069 & 46.016 \\
& AffectAgent-R \cite{ref20_affectagent} & 37.108 & 33.994 & 37.618 & 39.061 & 39.764 & 42.919 & 45.729 \\
& \textbf{APCF (ours)} & \textbf{43.095} & \textbf{35.265} & \textbf{39.450} & \textbf{43.940} & \textbf{45.112} & \textbf{44.685} & \textbf{46.708} \\
\bottomrule
\end{tabular}
\end{table*}

\begin{table*}[t]
\centering
\small
\setlength{\tabcolsep}{3.5pt}
\caption{Comparison with missing-modality methods adapted to open-vocabulary emotion generation. Cells report the dataset-native score (\%): level-1 F1 on OV-MERD+ and Avg on MER-FG.}
\label{tab:controlled}
\begin{tabular}{lp{6cm}ccccccc}
\toprule
\multicolumn{2}{c}{} & \multicolumn{7}{c}{Modality view} \\
\cmidrule(lr){3-9}
Dataset & Method & A & V & T & AV & AT & VT & AVT \\
\midrule
\multirow{6}{*}{OV-MERD+}
& MulT \cite{tsaiMultimodalTransformerUnaligned2019} & 9.491 & 10.361 & 47.778 & 18.591 & 47.786 & 47.702 & 47.791 \\
& MMIN \cite{zhaoMissingModalityImagination2021} & 17.740 & 21.856 & 48.359 & 22.249 & 48.230 & 48.351 & 48.399 \\
& MPLMM \cite{guoMultimodalPromptLearning2024} & 49.852 & 49.981 & 48.989 & 55.724 & 53.447 & 54.660 & 56.148 \\
& ComP \cite{heCrossmodalPromptingBalanced2026} & 41.342 & 35.055 & 50.147 & 14.592 & 49.797 & 49.832 & 49.783 \\
& BALM-FCM \cite{nguyenBALMModelAgnosticFramework2026} & 45.144 & 40.286 & 46.771 & 55.580 & 53.719 & 48.441 & 54.402 \\
& \textbf{APCF (ours)} & \textbf{56.490} & \textbf{50.587} & \textbf{54.116} & \textbf{58.818} & \textbf{61.086} & \textbf{61.193} & \textbf{63.015} \\
\midrule
\multirow{6}{*}{MER-FG}
& MulT \cite{tsaiMultimodalTransformerUnaligned2019} & 7.371 & 9.227 & 34.280 & 13.948 & 34.824 & 34.887 & 34.884 \\
& MMIN \cite{zhaoMissingModalityImagination2021} & 15.561 & 20.368 & 34.433 & 21.295 & 35.424 & 35.539 & 35.457 \\
& MPLMM \cite{guoMultimodalPromptLearning2024} & 40.306 & 34.955 & 38.500 & 41.482 & 42.936 & 41.944 & 44.173 \\
& ComP \cite{heCrossmodalPromptingBalanced2026} & 39.781 & 32.763 & 38.826 & 13.223 & 38.938 & 39.002 & 38.812 \\
& BALM-FCM \cite{nguyenBALMModelAgnosticFramework2026} & 38.816 & 34.734 & 39.135 & 39.921 & 43.150 & 40.714 & 43.828 \\
& \textbf{APCF (ours)} & \textbf{43.095} & \textbf{35.265} & \textbf{39.450} & \textbf{43.940} & \textbf{45.112} & \textbf{44.685} & \textbf{46.708} \\
\bottomrule
\end{tabular}
\end{table*}

\section{Experiments}
We evaluate released open-vocabulary systems, controlled fusion methods, APCF component variants, and the learned prototype space, for proving our proposed framework can achieve effective emotion modelling and recognition within incomplete modality context.

\subsection{Datasets and Evaluation Metrics}
\paragraph{Data and views.}
IEMOCAP \cite{bussoIEMOCAPInteractiveEmotional2008} supplies paired categorical and dimensional source supervision. All 31,327 MER-Caption+ examples \cite{lianAffectGPTNewDataset2025} are used for target training, and evaluation uses all 532 OV-MERD+ examples \cite{lianAffectGPTNewDataset2025} and all 1,200 MER-FG examples \cite{ref35_mer2025}. A fixed non-emotional sentence replaces unavailable native transcripts for 22 and five examples, respectively; these rows are not counted as controlled missing-text cases. Controlled views are instead produced by deleting complete streams. 
Every sample is evaluated under \(\{A,V,T,AV,AT,VT,AVT\}\) modality combinations.

\paragraph{Metrics.}
For both datasets, normalized terms are mapped through the same five emotion wheels. \(S_1\) and \(S_2\) are the mean F-scores over their coarse and fine mappings, and
\(\mathrm{Avg}=(S_1+S_2)/2\) is the primary metric. Avg is the native MER-FG ranking score and for OV-MERD+, its native score (\(S_1\) F-score) is used. All table entries are percentages; \(S_1\) and \(S_2\) remain available in the result artifacts. A frozen parser handles raw generations, and malformed or empty outputs remain failures.

\subsection{Experimental Setup}
Audio, video, and text are encoded by frozen Chinese-HuBERT-large \cite{hsuHuBERTSelfSupervisedSpeech2021}, CLIP ViT-L/14 \cite{ref40_clip}, and all-mpnet-base-v2 \cite{ref41_mpnet}, respectively. The decoder is Qwen2.5-7B-Instruct \cite{ref33_qwen25} with the released AffectGPT MER-Caption+ LoRA \cite{lianAffectGPTNewDataset2025,ref34_lora} merged and frozen. APCF uses eight prototypes and eight temporal slots, constructed from IEMOCAP dataset, a 384-dimensional evidence space, and a router at decoder layer 13. Generation is deterministic with at most 64 new tokens. All experiments are conducted on RTX 3090 GPUs.

Source initialization precedes eight epochs of target training. We employ a hybrid Muon--AdamW optimizer: Muon updates eligible two-dimensional hidden matrices with learning rate \(10^{-3}\), momentum \(0.95\), Nesterov momentum, and five Newton--Schulz iterations, while AdamW \cite{ref44_adamw} updates auxiliary parameters at \(5\times10^{-5}\) and prototype residuals at \(10^{-5}\). All parameter groups use weight decay \(0.01\). Learning rates follow a cosine schedule with 500 warm-up steps and decay to \(0.1\) of their initial values; the global gradient norm is clipped at \(1.0\). Controlled methods share the prompt, decoder, parser, and scorer; released systems retain their native front ends and prompting.

\subsection{Comparison with Open-Vocabulary Systems}
OV-MER \cite{lianOVMEROpenVocabularyMultimodal2025} uses its released acoustic and visual clue checkpoints with a view-aware merger. Emotion-LLaMA \cite{chengEmotionLLaMAMultimodalEmotion2024} is evaluated zero-shot from its released checkpoint, and AffectGPT \cite{lianAffectGPTNewDataset2025} retains its released multimodal modules without additional adaptation. AffectAgent \cite{ref20_affectagent} proposed module for incomplete modal fusion and was evaluated on open-vocabulary tasks, however, as it provides neither executable code nor a checkpoint, \emph{AffectAgent-R} denotes our paper-guided implementation of its reported agent, retrieval, fusion, and optimization components.

Table~\ref{tab:released-systems} shows that APCF ranks first under every modality view on both datasets. It gains 8.264 points for \(A\), 3.316 for \(AV\), and 2.913 for \(AT\) on OV-MERD+ dataset, and 5.987, 4.879, and 5.348 points for \(A\), \(AV\), and \(AT\) on MER-FG dataset, respectively. Text-rich views are already strong for language-decoder-based systems because the transcript is close to the generated output space, and APCF's larger gains on \(A\), \(AV\), and \(AT\) provide more direct evidence that it can organize the surviving non-textual cues.

\subsection{Comparison with Incomplete-Modality Fusion Methods}
Under the same target data, modality encoders, language decoder, prompt, evidence budget, and scorer, we adapt MulT \cite{tsaiMultimodalTransformerUnaligned2019}, MMIN \cite{zhaoMissingModalityImagination2021}, MPLMM \cite{guoMultimodalPromptLearning2024}, ComP \cite{heCrossmodalPromptingBalanced2026}, and BALM-FCM \cite{nguyenBALMModelAgnosticFramework2026} to open-vocabulary generation.

\paragraph{Open-vocabulary adaptation.}
For MulT, MMIN, MPLMM, and ComP, we remove the native fixed-label prediction head and pass the fused hidden sequenceinro the frozen language decoder. For BALM specifically, we insert its Feature Calibration Module (FCM), which can be transferred without altering the task. Gradient Rebalancing Module is not included because its design depends hevaily on fixed-label classification.

As shown in Table \ref{tab:controlled}, APCF leads on open-vocabulary emotion phrases generation, attributing its advantage more directly to incomplete-evidence organization under a common generator. The shared language decoder makes text-containing views a necessary but insufficient test of fusion. For MulT, MMIN, and ComP, their text-absent profiles are substantially weaker or unstable, particularly on \(V\) and \(AV\). Adding audio or video therefore contributes little once text is present; the apparently competitive text-view results mainly reflect transcript processing by the common language backbone. This also explains why comparing only \(AVT\) would substantially overestimate the missing-modality capability of these methods.

\subsection{Ablation Studies}
Table~\ref{tab:component-ablation} evaluates missing-view learning (MVL), source prototype transfer (SPT), prototype-conditioned sensory fusion (PSF), target prototype adaptation (TPA), and semantic causal routing (SCR). The upper block adds these components to a common base in order, while the lower block removes one component from full APCF. All variants retain the same backbone, dimensions, data, and evaluation protocol.

\begin{table}[t]
\centering
\small
\setlength{\tabcolsep}{2pt}
\caption{APCF component ablation. Filled and open circles denote enabled and disabled components. Values are the average over all seven modality combinations (All7, \%), using level-1 F1 on OV-MERD+ and Avg on MER-FG. Best results are bold.}
\label{tab:component-ablation}
\begin{tabular}{@{}*{5}{c}@{\hspace{4pt}}cc@{}}
\toprule
\multicolumn{5}{c}{Tested components} & \multicolumn{2}{c}{All7 (\%)} \\
\cmidrule(lr){1-5}\cmidrule(l){6-7}
MVL & SPT & PSF & TPA & SCR & OV-MERD+ & MER-FG \\
\midrule
\(\circ\) & \(\circ\) & \(\circ\) & \(\circ\) & \(\circ\) & 48.38 & 36.87 \\
\(\bullet\) & \(\circ\) & \(\circ\) & \(\circ\) & \(\circ\) & 53.87 & 40.36 \\
\(\bullet\) & \(\bullet\) & \(\circ\) & \(\circ\) & \(\circ\) & 53.42 & 40.93 \\
\(\bullet\) & \(\bullet\) & \(\bullet\) & \(\circ\) & \(\circ\) & 54.19 & 41.05 \\
\(\bullet\) & \(\bullet\) & \(\bullet\) & \(\bullet\) & \(\circ\) & 55.82 & 41.08 \\
\midrule
\(\circ\) & \(\bullet\) & \(\bullet\) & \(\bullet\) & \(\bullet\) & 47.69 & 37.36 \\
\(\bullet\) & \(\circ\) & \(\bullet\) & \(\bullet\) & \(\bullet\) & 55.05 & 42.18 \\
\(\bullet\) & \(\bullet\) & \(\circ\) & \(\bullet\) & \(\bullet\) & 53.94 & 41.96 \\
\(\bullet\) & \(\bullet\) & \(\bullet\) & \(\circ\) & \(\bullet\) & 53.89 & 41.84 \\
\(\bullet\) & \(\bullet\) & \(\bullet\) & \(\bullet\) & \(\bullet\) & \textbf{57.90} & \textbf{42.61} \\
\bottomrule
\end{tabular}
\end{table}

MVL produces the largest cumulative improvement, adding 5.49 points on OV-MERD+ and 3.49 on MER-FG; removing it from full APCF reduces performance by 10.21 and 5.25 points, respectively. The prototype components are also beneficial in the full system. Removing SPT, PSF, or TPA costs 2.85/0.43, 3.96/0.65, and 4.01/0.77 points on OV-MERD+/MER-FG, while adding SCR to the otherwise complete model contributes 2.08/1.53 points.
\subsection{Affect-Prototype Adaptation Analysis}

\begin{figure}[t]
\centering
\begin{subfigure}[t]{\columnwidth}
    \centering
    \includegraphics[width=.7\linewidth]{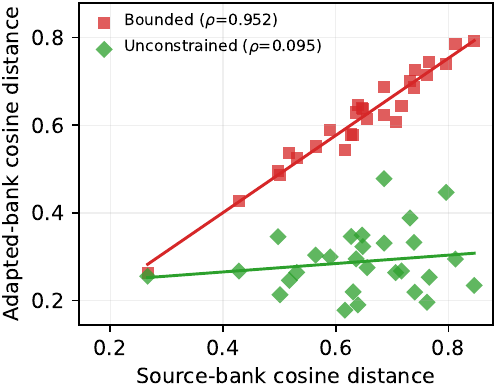}
    \caption{Pairwise source and adapted anchor distances.}
    \label{fig:prototype-source-geometry}
\end{subfigure}
\begin{subfigure}[t]{\columnwidth}
    \centering
    \includegraphics[width=.75\linewidth]{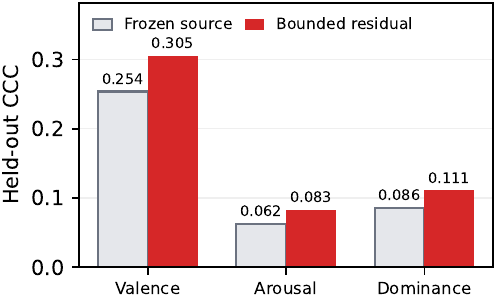}
    \caption{Axis-wise CCC on held-out emotion concepts.}
    \label{fig:prototype-vad-axis-ccc}
\end{subfigure}
\caption{Affect-prototype adaptation diagnostics averaged over target-training seeds 64, 65, and 71.}
\label{fig:prototype-analysis}
\end{figure}

Figure~\ref{fig:prototype-analysis} analyzes the two effects of target adaptation that are central to our prototype design. The bounded bank closely preserves the IEMOCAP pairwise anchor geometry (\(\rho=.952\)), whereas unconstrained adaptation largely destroys it (\(\rho=.095\)). On 1,024 held-out NRC-VAD concepts \cite{ref45_nrc_vad}, bounded adaptation also raises CCC \cite{ref48_ccc} from .254/.062/.086 to .305/.083/.111 for valence/arousal/dominance. These results show that bounded adaptation improves target-domain affect alignment while retaining the source structure.

\section{Conclusion}
This paper presents APCF, a candidate-free framework for open-vocabulary emotion recognition when any subset of audio, video, and text may be unavailable. APCF transfers a categorical--dimensional affect geometry from existing datasets, adapts it to target emotion phrases, uses the resulting prototypes to query and fuse only observed evidence, and routes that evidence causally into a frozen language decoder. In this way, emotion semantics influence the sensory fusion process rather than serving only as output labels. Comparisons on OV-MERD+ and MER-FG datasets with released open-vocabulary systems and controlled incomplete-modality pipelines show that APCF improves robustness across missing-input views while preserving competitive AVT performance.

\bibliography{Zotero}


\end{document}